\documentclass{article}
\usepackage{ijcai24}

\usepackage{times}
\usepackage{soul}
\usepackage{url}
\usepackage[hidelinks]{hyperref}
\usepackage[utf8]{inputenc}
\usepackage[small]{caption}
\usepackage{graphicx}
\usepackage{amsmath}
\usepackage{amsthm}
\usepackage{booktabs}
\usepackage{algorithm}
\usepackage{algorithmic}
\usepackage[switch]{lineno}
\usepackage{amssymb}
\usepackage{xspace}
\usepackage{xcolor} 
\usepackage{multirow}

\newcommand{\ie}{{\emph{i.e.}}\xspace}
\newcommand{\eg}{{\emph{e.g.}}\xspace}

\newcommand{\etal}{{\emph{et al.}}}

\title{Pick-and-Draw: Training-free Semantic Guidance for Text-to-Image Personalization }

\author{
Henglei Lv
\and
Jiayu Xiao\and
Liang Li\And
Qingming Huang\\
\emails
henglei.lv@vipl.ict.ac.cn
}

\begin{document}

\maketitle

\begin{abstract}
Diffusion-based text-to-image personalization have achieved great success in generating subjects specified by users among various contexts. Even though, existing finetuning-based methods still suffer from model overfitting, which greatly harms the generative diversity, especially when given subject images are few. To this end, we propose Pick-and-Draw, a training-free semantic guidance approach to boost identity consistency and  generative diversity for personalization methods. Our approach consists of two components: appearance picking guidance and layout drawing guidance. As for the former, we construct an appearance palette with visual features from the reference image, where we pick local patterns for generating the specified subject with consistent identity. As for layout drawing, we outline the subject's contour by referring to a generative template from the vanilla diffusion model, and inherit the strong image prior to synthesize diverse contexts according to different text conditions. The proposed approach can be applied to any personalized diffusion models and requires as few as a single reference image. Qualitative and quantitative experiments show that Pick-and-Draw consistently improves identity consistency and generative diversity, pushing the trade-off between subject fidelity and image-text fidelity to a new Pareto frontier.

\end{abstract}

\begin{figure}[ht]
    \vspace{-4mm}
    \begin{center}
        \includegraphics[width=1.0\linewidth]{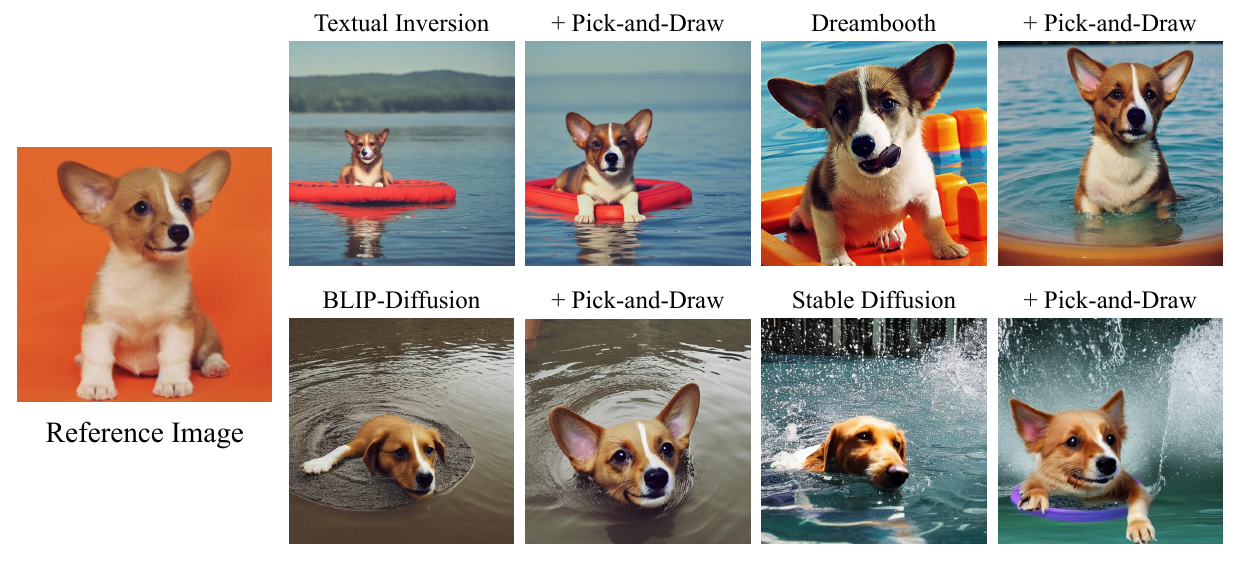}
    \end{center}
    \vspace{-3mm}
    \caption{Given a single reference image, Pick-and-Draw consistently improves identity consistency and image-text alignment over various personalization methods, including Textual Inversion, DreamBooth, and BLIP-Diffusion. The text prompt is ``A photo of a dog in water". Additionally, Directly applying Pick-and-Draw on vanilla Stable Diffusion also produces acceptable outcomes.}
    \label{fig:small_intro}
    \vspace{-3mm}
\end{figure}

\section{Introduction}

Recent large-scale diffusion models~\cite{dhariwal2021diffusion,ramesh2022hierarchical,rombach2022high,saharia2022photorealistic} demonstrate remarkable capability on text-to-image generation. Trained on billions of image-text pairs collected from the Internet, these models are competent to synthesize high-quality and diverse images conditioned on textual inputs. Owing to unprecedentedly strong image priors, text-to-image diffusion models are successfully applied to various downstream tasks, including image editing~\cite{hertz2022prompt,cao2023masactrl,epstein2023diffusion}, inpainting~\cite{yang2023paint,zhang2023towards}, augmentation~\cite{trabucco2023effective,fang2024data}, style transfer~\cite{yang2023zero,wang2023stylediffusion} and controllable generation~\cite{zhang2023adding,xiao2023r}. 

As a newly emerged task, text-to-image personalization aims to reason over specified subjects in assorted contexts. It requires the model to mimic the appearance of a subject given a reference image set, and synthesize the same subject in different contexts. Many works~\cite{gal2022image,ruiz2023dreambooth,li2023blip} are proposed and have achieved impressive results. However, these methods still suffer from severe mode collapse in data-scarce scenarios where only few reference images are available, and the diffusion network tends to simply memorize the few reference samples during the fine-tuning process. As a result, the model struggles to follow text instructions and synthesize subjects of different views, poses and backgrounds. To mitigate this problem, some works~\cite{ruiz2023dreambooth,kumari2023multi} leverage a regularization set to preserve the image priors of the original diffusion model, others propose to fine-tune a subset of model parameters~\cite{kumari2023multi} or introduce extra low-rank adaptors~\cite{hu2021lora}. These approaches help preserve the innate capabilities of the model, yet require extensive empirical hyperparameter tuning to obtain delicate results, and optimal hyperparameter configurations may vary across different subjects. Balancing the identity consistency and context diversity of generative outcomes remains a challenging problem.

To this end, we propose Pick-and-Draw, a training-free semantic guidance on text-to-image personalization methodologies, aiming to boost identity consistency while maintaining ability of diverse context synthesis. In general, our approach consists of two components: appearance picking guidance and layout drawing guidance. (1) As for appearance picking guidance, we feed the inverted latent of the reference image into the deep generative network, and extract a visual feature set as a palette, from which we pick ``color" for generating the specified subject. Specifically, we first adopt the layer-wise cross-attention maps corresponding to the specific subject, and threshold them to obtain salient binary masks. We then leverage the masks to extract the feature vectors within the object regions. Subsequently, we minimize the Unidirectional Relaxed Earth Mover Distance (UREMD) between the above feature vectors of the reference image and the generated image at each denoising step, so as to aid the model to better capture the appearance cues of the new concept during generation process.
(2) As for layout drawing guidance, we borrow the subject's shape and contour generated by the powerful original diffusion model as a template, and imitate the outline to enable diverse posture and context synthesis of the new concept for personalized model. To specify, we perform cross attention layout guidance to inject the shape and localization information to the personalized generation process. This helps align the generative contour with the template consistently during the denoising process, thereby inheriting the generative priors of the original model and ensuring diversity of the generated outcomes. The overall pipeline of our approach bears resemblance to the painting process of picking colors from a palette, drawing outlines based on a template, and subsequently applying colors to finalize the entire painting. In this sense, we term our method Pick-and-Draw.

Pick-and-Draw is a training-free plug-and-play semantic guidance approach developed for boosting text-to-image personalization, applicable to various personalized models including Texual Inversion~\cite{gal2022image}, DreamBooth~\cite{ruiz2023dreambooth}, and BLIP-Diffusion~\cite{li2023blip}, \etal\ Our method consistently improves personalized methods' identity consistency and generative diversity, pushing the trade-off between image fidelity and textual alignment to a new Pareto frontier. Moreover, we surprisingly find that directly applying Pick-and-Draw to vanilla Stable Diffusion~\cite{rombach2022high} also yields favorable outcomes.

To summarize, we make the following key contributions:
\begin{enumerate}
\item We propose Pick-and-Draw, a training-free semantic guidance approach to enhance identity consistency and generative diversity for text-to-image personalization models.
\item We demonstrate quantitatively and qualitatively that Pick-and-Draw consistently improves identity preservation and diverse context synthesis of various personalized models, pushing the trade-off between subject fidelity and image-text fidelity to a new Pareto frontier.
\item We find that directly applying Pick-and-Draw to vanilla Stable Diffusion yields surprisingly favorable outcomes, which may potentially inspire research on training-free single-image personalization. 
\end{enumerate}

\section{Related Work}
\subsection{Text-to-image diffusion}
Diffusion models are a class of generative models that learn image distributions through sequentially denoising. A diffusion model consists of a diffusion process and a reverse process. Given an initial image $x_0$, the diffusion process gradually adds Gaussian noise $\epsilon_t$ in $T$ time-steps until $x_0$ is diffused into $x_T$ which conforms to a Gaussian distribution. The reverse process aims to recover $x_0$ given $x_T$ by training a denoiser $\epsilon_\theta$ that predicts the noise $\epsilon_t$ given timestep $t$ and the noisy image $x_t$ using diffusion loss:
\begin{equation}
    L(\theta) = \mathbb{E}_{x_0, t, \epsilon_t\sim\mathcal{N}(0,1)}([||\epsilon_t-\epsilon_\theta(x_t, t)||^2]).
\end{equation}

Stable Diffusion (abbreviated as SD)~\cite{rombach2022high} is a powerful text-conditioned latent diffusion model which performs diffusion in the latent space $Z$ instead of the pixel space $X$ and injects text condition into the diffusion process, allowing for flexible conditional generation.

\subsection{Energy functions in diffusion models}
From a score-based perspective, each step in the reverse process in a diffusion model can be seen as an estimate of a score function $\nabla_{x_t}\log p(z_t)$~\cite{song2020score}. Given external condition $y$, diffusion models generate conditional samples from $p(z_t|y)\propto p(z_t)p(y|z_t)$. The first term $p(z_t)$ corresponds to the unconditional score function, and the second term $p(y|z_t)$ is equivalent to an energy function $\mathcal{E}(z_t;t,y)$. Numerous energy functions have been proposed and used in various tasks, including classifier guidance~\cite{dhariwal2021diffusion}, CLIP scores~\cite{nichol2021glide} and penalties on attention~\cite{chen2024training,xie2023boxdiff,xiao2023r,epstein2023diffusion}. In this sense, We propose two energy functions to boost identity consistency and maintain context diversity respectively.

\subsection{Text-to-image personalization}
Personalization aims to reason over specified subjects in assorted contexts. Textual Inversion~\cite{gal2022image} learns an embedding of an unique word to represent the specified subject. DreamBooth~\cite{ruiz2023dreambooth} fine-tunes the whole diffusion UNet to bind a unique identifier with the specified subject. Custom Diffusion~\cite{kumari2023multi} fine-tunes the key and value projection matrices in the cross attention layers in the diffusion UNet. Encoder-based methods~\cite{wei2023elite,jia2023taming,gal2023designing} fine-tunes the text encoder and potentially trains an image encoder or multi-modal encoder~\cite{li2023blip} to encode example images of specified subjects into embedding that is leveraged in personalized generation. Our proposed Pick-and-Draw consistently improves identity reproduction of these personalization methods without harming the model diversity in a training-free one-shot manner.

\begin{figure*}[ht]
    \vspace{-6mm}
    \begin{center}
        \includegraphics[width=1.0\linewidth]{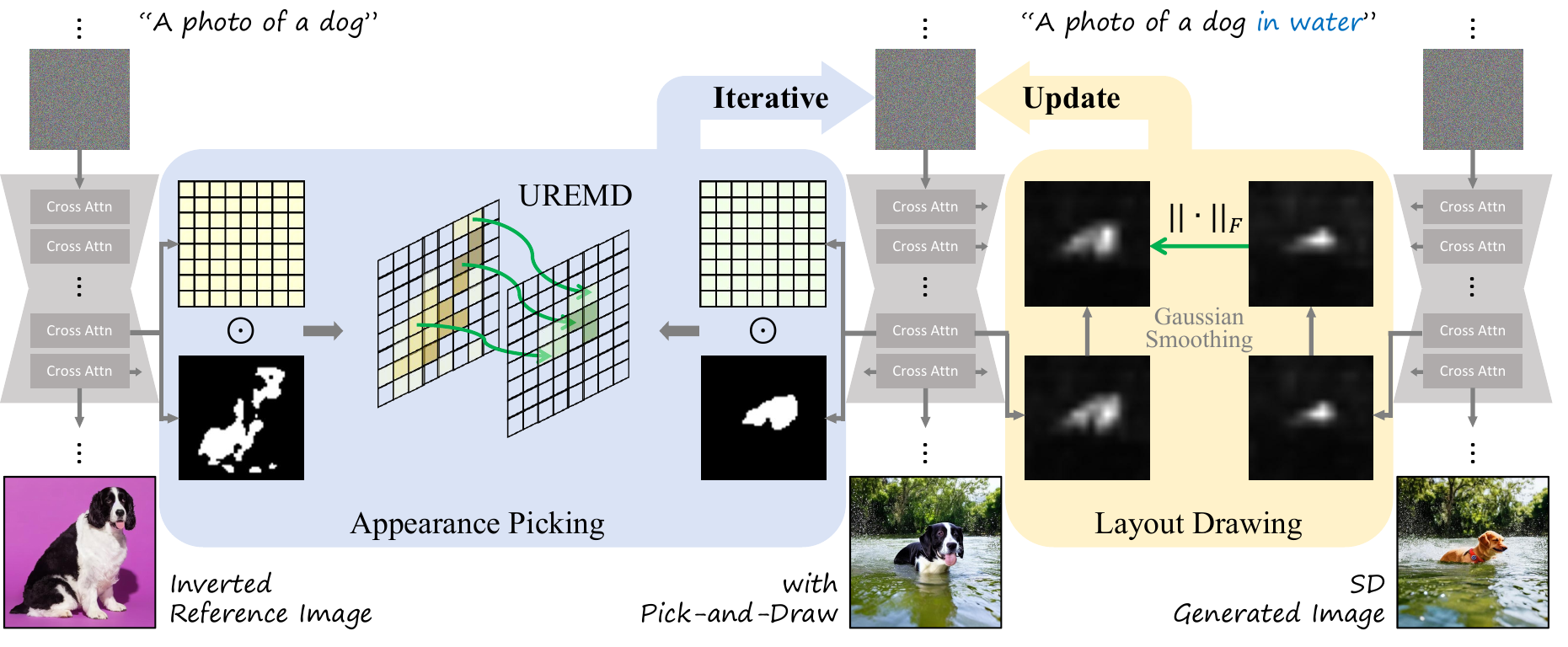}
    \end{center}
    \caption{Overall pipeline of our proposed Pick-and-Draw. We iteratively refine the generative outcomes via appearance picking and layout drawing, which is achieved by optimizing a designed score function. In appearance picking, we pick saliency-aware features from certain cross attention decoder layers, and transfer the appearance cues by minimizing the Unidirectional Relaxed Earth Movers Distance (UREMD), aiming to boost identity consistency. For layout drawing, we extract cross attention maps in every cross attention layer, smooth them with a Gaussian kernel, then minimize the Frobenius norm to draw the subject outline. This localizes the appearance transfer within the subject-relative regions  and introduces novel layout from the vanilla Stable Diffusion, so as to improve generative diversity.}
    \label{fig:introduction}
    \vspace{-3mm}
\end{figure*}

\section{Method}
Our purpose is to conduct training-free surgery on personalized diffusion models to boost identity consistency and generative diversity. Given a reference image $I_r$ depicting a specified subject $s$ (\eg, dog), we aim to generate a personalized picture $I_p$ with a text prompt $P_{p}$ (\eg ``A dog in water"), which is consistent with instance $I_r$ and conveys contextual semantics specified by $P_p$. 

Current personalized models fail to reconcile both identity consistency and generative diversity, since they are prone to overfit on the given few subject images, inevitably reducing the generative latent space of diffusion models to a lower dimension. To alleviate the above issues, our core idea is to inject appearance information of the reference image and contextual priors from the original diffusion into the personalized generative process. The overall architecture is shown in Fig.~\ref{fig:introduction}. Our pipeline simulates a human painting process, we (1) adopt the intermediate feature set of the reference image as a palette, where we pick ``colors" (\ie, representative feature vectors that convey appearance information) to blend the new subject on the canvas, and (2) draw the outlines based on a generative template. We develop an appearance picking guidance and a layout drawing guidance for the above two procedures respectively, and iteratively update the noisy latent via optimizing a designed energy function. The appearance picking guidance helps preserve the subject identity and the layout drawing guidance ensures the generative diversity. We first discuss cross attention saliency map extraction and selection strategy in Sec.~\ref{sec:ca_map}, then introduce our two types of semantic guidance in Sec.~\ref{sec:app_guidance} and Sec.~\ref{sec:layout_guidance} respectively.

\subsection{Cross Attention Map Extraction and Selection}
\label{sec:ca_map}
Both appearance guidance and layout guidance rely upon a saliency mask that highlights the subject-relative region. Previous works~\cite{hertz2022prompt,tumanyan2023plug,xiao2023r} show that cross attention maps contain rich semantic and layout information. Similarly, we extract the subject-relevant cross attention maps at each layer $l$ of the diffusion UNet:
\begin{equation}
\mathcal{A}_l=\text{softmax}(\frac{Q_lK_l^T}{\sqrt{d}}),
\end{equation}
where $Q_l$ is the query features projected from the image features, $K_l$ is the key features projected from the textual embedding with corresponding projection matrices, and $d$ is a scaling factor. We perform min-max normalization on these maps to acquire a set of normalized cross attention saliency maps $\hat{\mathcal{A}}_{0:L} = \{\hat{\mathcal{A}}_0,\hat{\mathcal{A}}_1,...,\hat{\mathcal{A}}_L\}$. We omit the timestep $t$ for simplicity.

\begin{figure}[t]
    \begin{center}
        \includegraphics[width=1.0\linewidth]{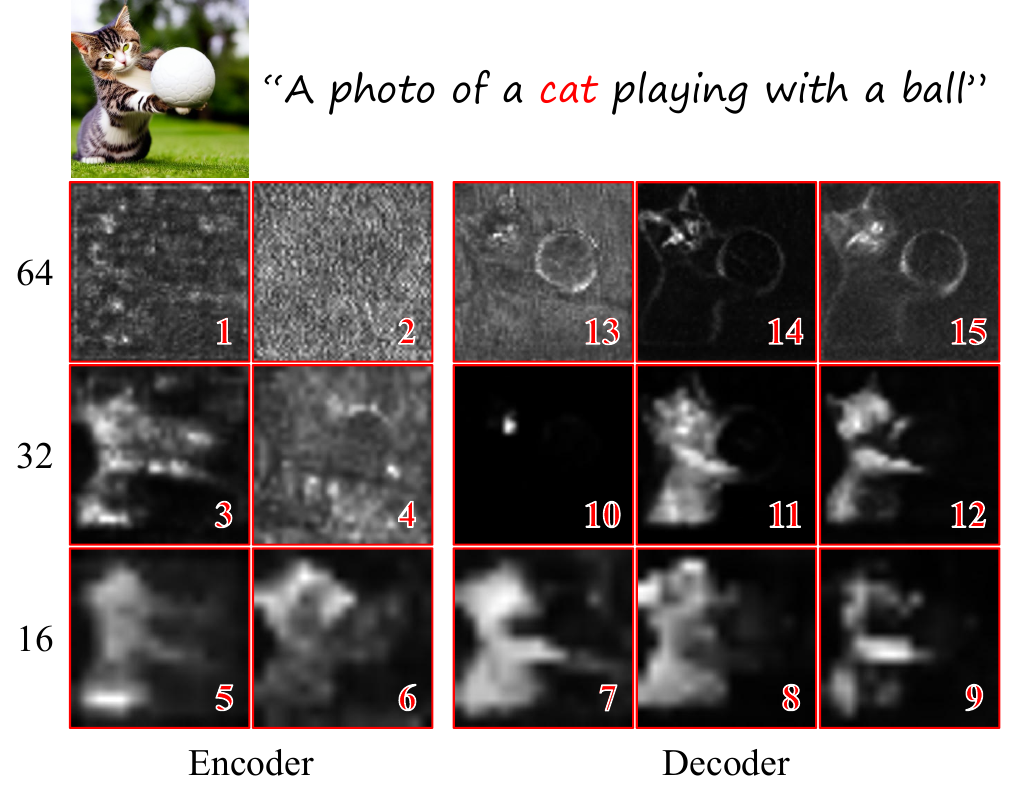}
    \end{center}
    \vspace{-5mm}
    \caption{Illustration of cross attention maps extracted from different layers in the encoder and decoder of the UNet, numbered by inference order. Resolution is marked on the left.}
    \label{fig:ca_maps}
\end{figure}

Cross attention maps of each layer contain different semantic information and highlight different regions of the image. Taking Stable Diffusion~\cite{rombach2022high} as an example, the UNet consists of multiple up-blocks and down-blocks, with each block containing several cross-attention layers. We visualize the cross attention maps corresponding to different layers in Fig.~\ref{fig:ca_maps}. Maps from deeper blocks have smaller resolutions. Specifically, we observe that (1) maps of 64$\times$64 resolution are fine-grained and tend to outline edges of all salient objects. They capture the high-frequency attributes, yet contain much background noise; (2) maps of 32$\times$32 resolution are better aligned with the subject and highlight different regions, entailing richer semantic information; (3) maps of 16$\times$16 resolution are coarse and well aligned, reflecting the approximate layout information of objects; (4) maps from encoder layers are mostly blended with more background noise, while maps from decoder layers better align with the subject layout, which convey richer semantic and structural information.

Due to the disparate layout granularity and semantic information of cross attention maps from each layers, we leverage different sets of maps for our proposed two types of semantic guidance. For appearance picking guidance, we need to select the most representative activations to provide appearance cues. Those whose corresponding attention maps align well to the subject (or part of it) are most desired. For layout drawing guidance, we need to introduce novel image layout from an external prior. Note that the image layout not only includes the subject contour, but also the context and the background, thereby we utilize attention maps from all layers.

\begin{figure*}[ht]
    \vspace{-6mm}
    \begin{center}
        \includegraphics[width=1.0\linewidth]{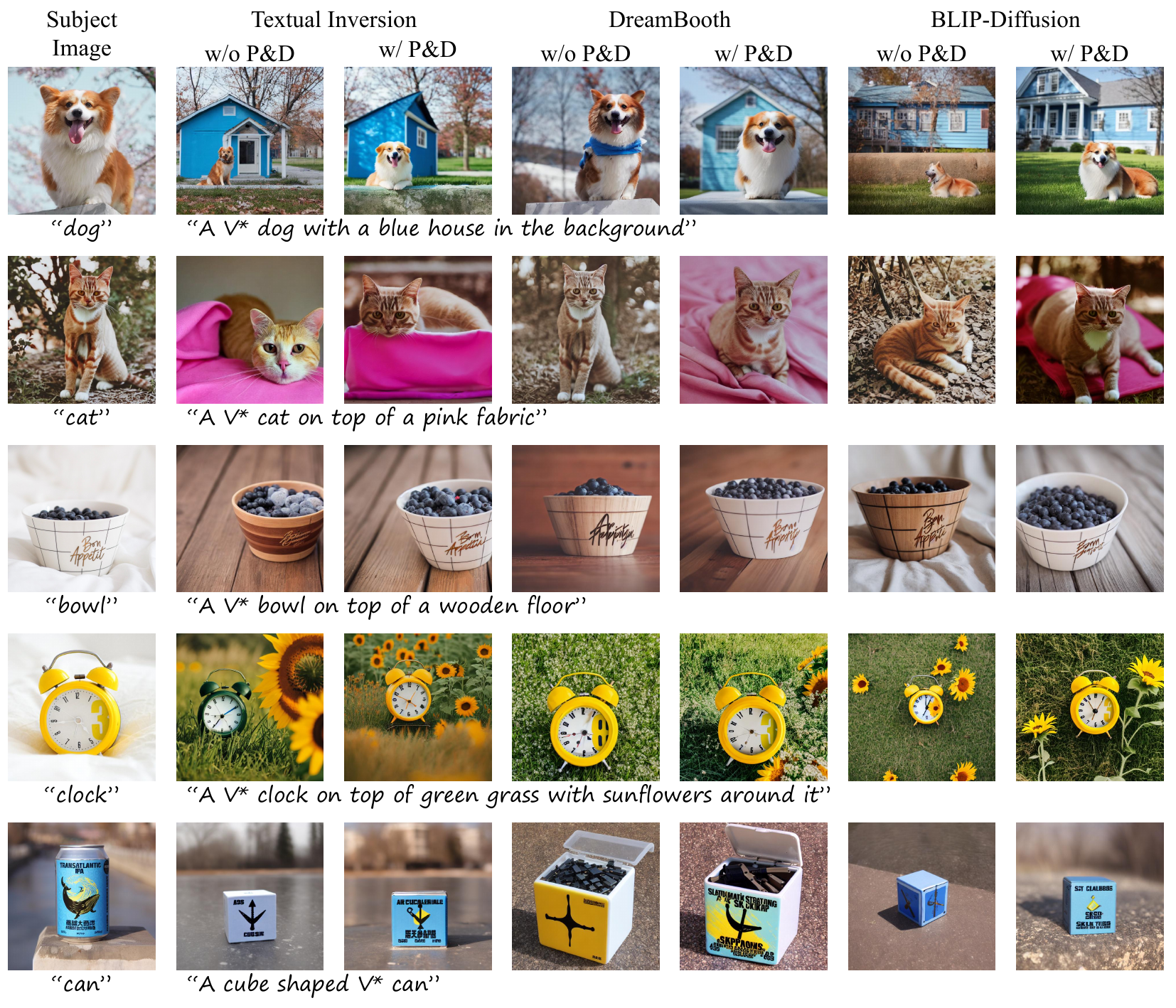}
    \end{center}
    \vspace{-3mm}
    \caption{Qualitative results on different baselines with and without Pick-and-Draw. The format of text prompt slightly differs across the three baselines and we choose the DreamBooth format for presentation.}
    \label{fig:qualitative}
    
\end{figure*}

\subsection{Appearance Picking Guidance}
\label{sec:app_guidance}

Intuitively, we regard the generative feature set associated with the object region of the reference image as a palette, which provides essential visual cues for object appearance. To generate subjects which are consistent with the reference image, we assign the closest element in the palette to each generative feature vector within the subject-related region, and minimize a transport distance to facilitate the diffusion model to gradually capture the appearance essence from the reference image during the generation process.

First, in order to obtain intermediate features of a reference image that depict the image content at each denoising step,  we invert the reference image to the initial random noisy latent, and feed it into diffusion model to reproduce the denoising trajectory. We adopt Null-Text Inversion~\cite{mokady2023null} as the inversion approach, which aligns the diffusion latent trajectory with the denoising trajectory by optimizing a null-text unconditional embedding in each step.

Second, we spot the local regions corresponding to the specific subject of reference image and generative image, and accordingly extract the saliency-aware feature sets for appearance transfer. Specifically, we apply hard masks $\mathcal{M}_l$ derived from normalized attention maps $\hat{\mathcal{A}}_l$  of each attention layer $l$ to highlight the subject-relative regions of the dense feature $\Psi_l$, and acquire saliency-aware visual features $\mathcal{V}_l$:

\begin{equation}
\begin{aligned}
   (\mathcal{M}_l)_{h,w} &= 
   \begin{cases}
        1 &\text{if } (\hat{\mathcal{A}}_l)_{h,w} \ge \tau \\
        0 &\text{if } (\hat{\mathcal{A}}_l)_{h,w} < \tau
    \end{cases},
    \\
    \mathcal{V}_l &= \mathcal{M}_l \odot \Psi_l,
\end{aligned}
\end{equation}
where $\tau$ is a threshold parameter and tuple $(h, w)$ represents a spatial entry of the attention map. We then extract the non-zero channels of $\mathcal{V}_l\in\mathbb{R}^{H_l\times W_l\times D_l}$ to obtain a feature set $\mathbf{V}_l=\{(V_l)_1, ..., (V_l)_n\}$, where $(V_l)_i$ is the $i$-th feature vector at the $l$-th layer and $n$ is the number of pixels within the salient region.  

Subsequently, we search for an effective way to transfer the appearance essence from the reference palette to the generative canvas during the denosing process. Previous works provide inspiration that leverage Earth Movers Distance (EMD) to model the divergence between two feature distributions.
Specifically, let $\mathbf{A}=\{A_1, ... ,A_n\}$ and $\mathbf{B}=\{B_1, ... ,B_m\}$ be two sets of $n$ and $m$ feature vectors, respectively. The EMD is formulated as:
\begin{equation}
\centering
\begin{aligned}
\textsc{EMD}(\mathbf{A},\mathbf{B})&=\min_{\mathbf{T}\geq 0} \sum_{ij} \mathbf{T}_{ij}\mathbf{C}_{ij},  \\
s.t. \sum_j &\mathbf{T}_{ij} = 1/m,  \\
\sum_i &\mathbf{T}_{ij} = 1/n,
\label{eq:emdb}
\end{aligned}
\end{equation}
where $\mathbf{T}$ is the transport matrix which defines partial pairwise
assignments, and $\mathbf{C}$ is the cost matrix which defines the distance between an
element in $\mathbf{A}$ and an element in $\mathbf{B}$. 

The EMD measures the cost of bidirectional optimal transport between two sets of features. In context of appearance transfer, we aim to minimize the unidirectional optimal transport cost from the generated image features $\mathbf{V}_l^{\text{gen}}=\{(V_l^{\text{gen}})_1, ..., (V_l^{\text{gen}})_m\}$  to the reference image features $\mathbf{V}_l^{\text{ref}}=\{(V_l^{\text{ref}})_1, ..., (V_l^{\text{ref}})_n\}$. We relax the EMD to single constraint and define the Unidirectional Relaxed Earth Movers Distance (UREMD) as the  appearance-aware loss:
\begin{equation}
\begin{aligned}
\ell_{\text{app}}&=\text{UREMD}(\mathbf{V}^{\text{ref}},\mathbf{V}^{\text{gen}})\\
&=\min_{\mathbf{T}\geq 0} \sum_{ij} \mathbf{T}_{ij}\mathbf{C}_{ij},\\
& s.t. \sum_i \mathbf{T}_{ij} = 1/n.
\end{aligned}
\end{equation}

We aim to assign the closest element in  $\mathbf{V}_l^{\text{ref}}$ to $V_j^{\text{gen}}$. In this manner, the aforementioned formulation is equivalent to:

\begin{gather}
    \ell_{\text{app}}\,=\,\frac{1}{n}\sum_j\min_i\mathbf{C}_{ij},
\end{gather}
where we define the $(i,j)$-th entry $\mathbf{C}_{ij}$ of cost matrix $\mathbf{C}$ as the pairwise cosine distance between two feature vectors:
\begin{equation}
\mathbf{C}_{ij}\,=\,D_{\cos}(V^{\text{ref}}_i,V^{\text{gen}}_j)\,=\,1-\frac{V^{\text{ref}}_i\cdot V^{\text{gen}}_j}{ \|V^{\text{ref}}_i\|\|V^{\text{gen}}_j\|}.
\end{equation}

Simply put, at each step we find a one-to-one injection from the generated features to the reference features, and the mean UREMD between these two feature sets can be considered a metric evaluating overall subject appearance similarity. Optimizing the appearance-aware loss helps align the feature distributions of the reference image and generated image while avoiding excessive constraints on the generative layout, which is crucial for preserving the quality of the generated outcomes.

\subsection{Layout Drawing Guidance}
\label{sec:layout_guidance}
Previous works have successfully utilized cross attention layout control on image editing~\cite{hertz2022prompt,epstein2023diffusion} and grounded generation~\cite{chen2024training,xie2023boxdiff,xiao2023r}. The key idea is that cross attention maps highlight the salient object-related region, specifying shape, posture and position of objects within the canvas. Since vanilla Stable Diffusion which is trained on massive image-text pairing datasets has been proven of impressive generative diversity, we aim to perform layout guidance to borrow its image prior and guide personalized generation. We regard the subject’s contour generated by vanilla Stable Diffusion as a template and draw the outline by imitating. In this way, the personalized model inherit strong generative priors of SD, ensuring the diversity of generative outcomes. 

Unlike most works which aggregate the attention maps from each layers to a single saliency map, we leverage the cross attention maps in all layers separately, as discussed in Sec.~\ref{sec:ca_map}. Following Chefer \etal ~\shortcite{chefer2023attend}, we first apply a Gaussian kernel on these maps to obtain smoothed attention maps, aiming to eliminate noisy perturbations, then calculate the distance of layer-wise attention maps between the generated image and the template image to as layout-aware loss:
\begin{equation}
    \ell_{\text{lay}}=\frac{1}{L}\sum_l\|G(\hat{\mathcal{A}}_l^{\text{gen}})-G(\hat{\mathcal{A}}_l^{\text{temp}})\|_F,
\end{equation}
where $G$ is the Gaussian kernel, $\|\cdot\|_F$ is the Frobenius norm and $L$ is the number of cross-attention layers.

By aligning the generative layout between the personalized model and the original diffusion model, we help the model inherit the strong generative priors, so as to synthesize diverse context according to different text conditions, and thus alleviate the model overfitting problem during generation. In practice, we also find that applying the layout-aware loss in early steps of the denoising process helps pre-stablizing the subject's contour, providing a good initialization for performing the appearance guidance. 

By combining the appearance picking guidance and layout drawing guidance together, we generate new subjects that highly align with the reference image among various contexts, pushing the trade-off between identity consistency and generative diversity to a new Pareto frontier. The overall loss function at step $t$ for personalized generation can be written as below:
\begin{equation}
    \ell_t = \alpha_t\ell_{app} + \beta_t\ell_{lay},
\end{equation}
and the noisy latent $z_t$ is iteratively updated at step $t$ by
\begin{equation}
    z_t\leftarrow z_t-\eta_g\nabla_{z_t}l_t,
\end{equation}
where $\eta_g$ is the guidance ratio.

\section{Experiments}
\subsection{Main Qualitative Results}
We provide qualitative comparisons of various text-to-image personalization methods including Textual Inversion~\cite{gal2022image}, DreamBooth~\cite{ruiz2023dreambooth} and BLIP-Diffusion~\cite{li2023blip} with and without Pick-and-Draw in Fig.~\ref{fig:qualitative}. See Appendix for detailed description of the baselines. We observe that Pick-and-Draw consistently improves both subject-fidelity and image-text fidelity on all three baselines. Textual Inversion falls short in preserving identity; DreamBooth and BLIP-Diffusion better preserve the subject identity, but tend to overfit and memorizes the subject's pose and background, causing unsatisfactory alignment with the text prompt. Specifically, DreamBooth fails to generate the blue house ($1\textsuperscript{st}$ row) and sunflowers ($4\textsuperscript{th}$ row) and overfits to the cat image ($2\textsuperscript{nd}$ row), while BLIP-Diffusion memorizes background of the forest ($2\textsuperscript{nd}$ row) and the white fabric ($3\textsuperscript{rd}$ row). All three methods fail to preserve appearance traits when performing geometric shape modification ($5\textsuperscript{th}$ row). Furthermore, many synthesized subjects exhibit inconsistencies in certain details.

In comparison, our proposed Pick-and-Draw (1) greatly enhances identity consistency, (2) improves generative diversity and aligns better with the text prompt. The effectiveness of Pick-and-Draw can be attributed to the accurate semantics provided by appearance picking guidance and layout drawing guidance. The appearance picking guidance performs vigorous relaxed appearance transfer by calibrating the misalignment between the generated and reference appearance cues sets while the layout drawing guidance forces the personalized generation to an external layout, thereby mitigates the overfitting problem.

\begin{table}[tb] 
    \begin{center}
    \renewcommand\arraystretch{1.1}
    \resizebox*{1\linewidth}{!}{
        \begin{tabular}	{l c  l l l}
        \toprule
        \midrule
        {\textbf{Methods}} &\textbf{P\&D} & \textbf{DINO} & \textbf{CLIP-I} & \textbf{CLIP-T} \\
        \cmidrule(lr){1-5}
        \multicolumn{2}{l}{Real Images (Oracle)}          & 0.774 & 0.885 & \quad-\\
        \cmidrule(lr){1-5}
        \multirow{2}{*}{Stable Diffusion}    &$\times$       & 0.320 & 0.504 & 0.339 \\
                            &$\checkmark$   & 0.552\tiny{\textcolor{blue}{\ (+0.232)}} & 0.641\tiny{\textcolor{blue}{\ (+0.137)}} & 0.335\tiny{\textcolor{red}{\ (-0.004)}} \\
        \cmidrule(lr){1-5}
        \multirow{2}{*}{Textual Inversion}   &$\times$       & 0.568 & 0.664 & 0.252 \\
                            &$\checkmark$   & 0.627\tiny{\textcolor{blue}{\ (+0.059)}} & 0.745\tiny{\textcolor{blue}{\ (+0.081)}} & 0.263\tiny{\textcolor{blue}{\ (+0.011)}} \\
        \cmidrule(lr){1-5}
        \multirow{2}{*}{BLIP-Diffusion}      &$\times$       & 0.587 & 0.716 & 0.292 \\
                            &$\checkmark$   & 0.651\tiny{\textcolor{blue}{\ (+0.064)}} & 0.778\tiny{\textcolor{blue}{\ (+0.062)}} & 0.300\tiny{\textcolor{blue}{\ (+0.008)}} \\
        \cmidrule(lr){1-5}
        \multirow{2}{*}{DreamBooth}          &$\times$       & 0.616 & 0.739 & 0.297 \\
                            &$\checkmark$   & 0.696\tiny{\textcolor{blue}{\ (+0.080)}} & 0.790\tiny{\textcolor{blue}{\ (+0.051)}} & 0.303\tiny{\textcolor{blue}{\ (+0.006)}} \\

        
        \bottomrule
        
        \end{tabular}
    }
    \end{center}
\caption{Quantitative comparisons on DreamBench dataset. P\&D are short for our proposed method Pick-and-Draw. Performance gains and losses are written in blue and red subscripts, respectively.}
\label{tab::quantitative}
\end{table}

\begin{figure}[h]
    \vspace{-7mm}
    \begin{center}
        \includegraphics[width=1.1\linewidth]{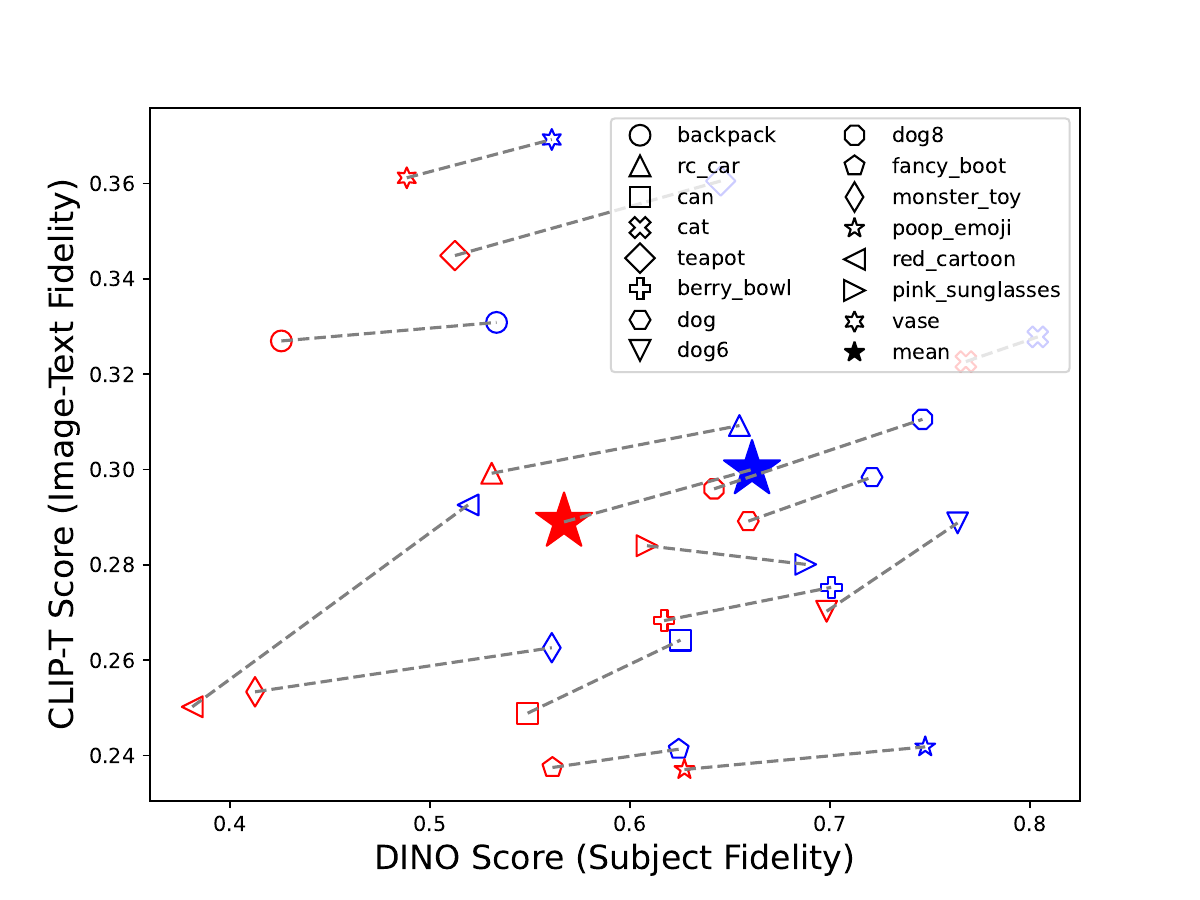}
    \end{center}
    \vspace{-4mm}
    \caption{Alignment metrics of BLIP-Diffusion before (red) and after (blue) applying Pick-and-Draw for sample subjects.}
    \vspace{-2mm}
    \label{fig:tradeoff}
\end{figure}

\subsection{Comparisons on DreamBench Dataset}
In Tab.~\ref{tab::quantitative}, we reproduce three personalization methods and study the impact of Pick-and-Draw quantitatively on DreamBench dataset~\cite{ruiz2023dreambooth} , containing 30 subjects and 25 text prompts for each subject. Particularly, we include vanilla Stable Diffusion as an additional baseline, and report DINO and CLIP-I scores of real images as the performance upper bound. Following DreamBooth~\cite{ruiz2023dreambooth}, we report DINO~\cite{caron2021emerging}, CLIP-I~\cite{radford2021learning} and CLIP-T scores. DINO and CLIP-I scores evaluate subject fidelity and CLIP-T scores evaluate image-text fidelity (see appendix for detailed description of the metrics). For every text prompt, we generate 4 images, summing up to a total of 3,000 images across all subjects. 

The overall results are consistent to the qualitative findings, where Pick-and-Draw improves the performance of the three methods on all metrics. The remarkable performance gains on DINO and CLIP-I metrics can be attributed to the appearance picking guidance which performs accurate relaxed appearance transfer and greatly enhances identity consistency. The improved CLIP-T score can be attributed to the layout drawing guidance, which anchors the generated layout to an external prior and introduces more generative diversity, leading to better alignment with the text prompt. Additionally, in Fig.~\ref{fig:tradeoff} we show per-subject metrics and observe that Pick-and-Draw significantly improves subject fidelity and while considerably improves image-text fidelity in most cases.

\begin{figure}[ht]
    \vspace{-2mm}
    \begin{center}
        \includegraphics[width=1.0\linewidth]{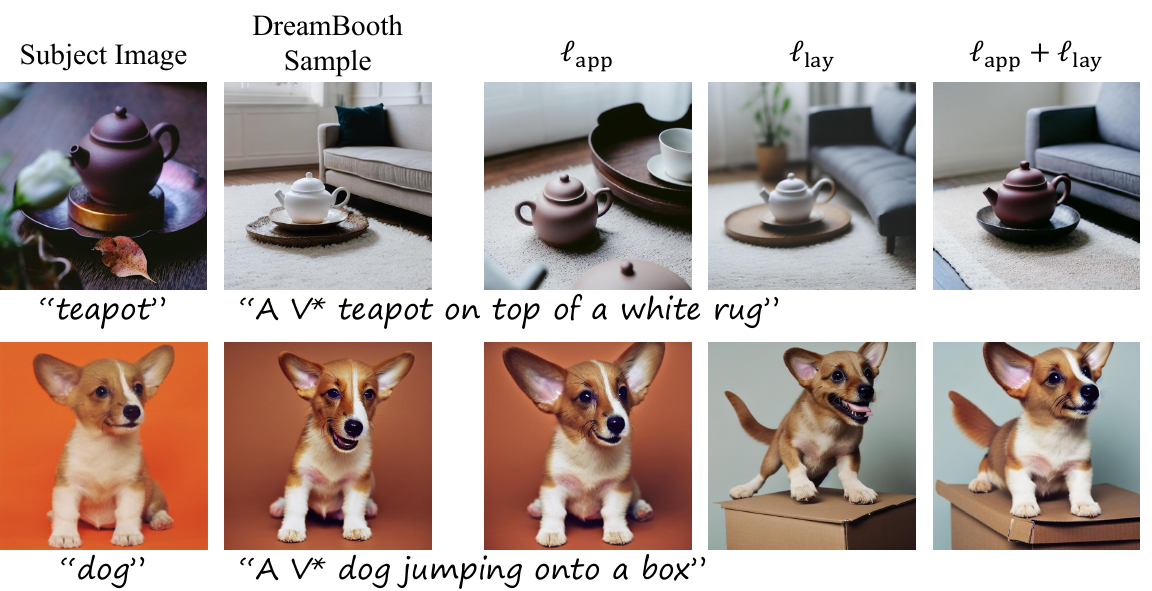}
    \end{center}
    \vspace{-2mm}
    \caption{Ablation study for the effect of different losses, including the appearance-aware loss $\ell_{\text{app}}$, layout-aware loss $\ell_{\text{lay}}$ and both combined, conducted on DreamBooth.}
    \label{fig:ablation_losses}
\end{figure}

\begin{table}[tb] 
    \begin{center}
    \renewcommand\arraystretch{0.9}
    \resizebox*{1\linewidth}{!}{
        \begin{tabular}	{l c  c c c c c c}
        \toprule
        \midrule
           & Textual Inversion&$16\times16$ &\quad & $32\times32$ &\quad & $64\times64$  \\
        \cmidrule(lr){1-7}
        \textbf{DINO}  & 0.568&0.582&\quad  & \textbf{0.627}&\quad  & 0.593    \\
        \cmidrule(lr){1-7}
        \textbf{CLIP-I}    & 0.664& 0.709&\quad  & \textbf{0.745}&\quad  & 0.718   \\
        \cmidrule(lr){1-7}
        \textbf{CLIP-T}  & 0.252 & 0.241&\quad  &\textbf{0.263} &\quad  & 0.259   \\
        \bottomrule
        \end{tabular}
    }
    \end{center}
    \caption{Ablation results on different activation selection strategy for the appearance picking guidance, conducted on Textual Inversion. Best results are bold.}
    \label{tab::ablation_activation}
\end{table}

\begin{figure}[t]
    \begin{center}
        \includegraphics[width=1.0\linewidth]{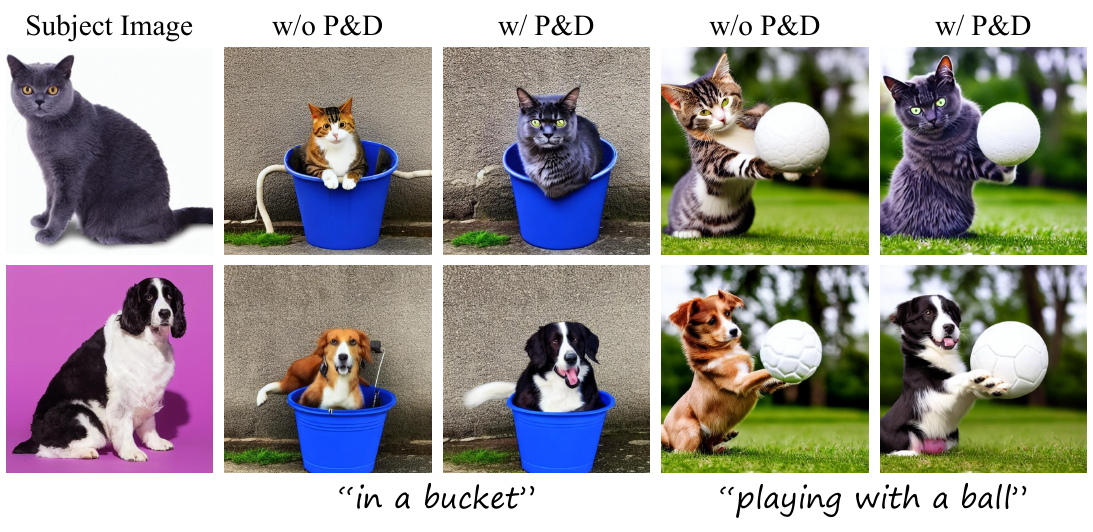}
    \end{center}
    \vspace{-2mm}
    \caption{Visual results of Pick-and-Draw directly applying to vanilla Stable Diffusion.}
    \label{fig:sd_results}
\end{figure}

\subsection{Ablation Study}

\noindent\textbf{Impact of activation selection.} In Tab.~\ref{tab::ablation_activation}, we quantitatively study the impacts of different activation selection strategy for the appearance picking guidance discussed in  Sec.~\ref{sec:app_guidance} for ablation. We choose Textual Inversion as the example baseline. We find that using activations of resolution $32\times 32$ yields the best results on both subject fidelity and image-text fidelity. Specifically, activations of resolution $16\times16$ contain mainly layout information, thus unwanted layout leakage might happen and reduces the text prompt alignment. Activations of resolution $64\times64$ focus on fine-grained high-frequency details such as edges, which are not sufficient for local appearance transfer. In comparison, activations of resolution $32\times32$ encode rich semantic information and focus on different regions of the subject, facilitating the local appearance transfer and achieve the overall best result. This observation is consistent with the discussion in Sec.~\ref{sec:ca_map}. Qualitative results are provided in Appendix for intuitive visualization.
\\

\noindent\textbf{Impact of two loss components.} We show visual results for comprehension of our proposed appearance-aware loss $\ell_{\text{app}}$ and  layout-aware loss $\ell_{\text{lay}}$ in Fig.~\ref{fig:ablation_losses}. We present two failure cases of DreamBooth, \ie the attribute misbinding issue in the first row and the overfitting issue in the second row, and illustrate how the two losses work together to address these issues collectively. The $\ell_{\text{app}}$ facilitates local appearance transfer, which make the generated subjects (the $3$-rd column) more aligned with the reference image. However, the appearance transfer may be unbounded and incorrect (the $1$-st row) and novel layout are not introduced (the $2$-nd row). The $\ell_{\text{lay}}$ (the $4$-th column) constrains the appearance transfer within the subject region (the $1$-st row) and introduces novel layout (the $2$-nd row), but without appearance picking guidance, the identity consistency of the new subject is not guaranteed. When combining the two losses together for diffusion guidance, the generated outcomes (the last column) exhibit the best performance. Specifically, the appearance transfer accurately bounded within the subject region, solving the attribute misbinding issue, while novel layout is introduced with consistent subject identity, mitigating the overfitting problem.

We also discuss the impact of different guidance timestep selection. Quantitative results and analysis are provided in Appendix.

\subsection{Results on Vanilla Stable Diffusion}
We directly apply Pick-and-Draw on vanilla Stable Diffusion and observe surprisingly favorable outcomes in some cases. Visual results are presented in Fig.~\ref{fig:sd_results}. Without the strong subject prior of the fine-tuning based personalization baselines, the appearance transfer still ensures identity consistency when the SD generated subject and the reference subject share similar shape and size. The bottom right of Fig.~\ref{fig:sd_results} is a failure case, where the size of generated subject is inconsistent with that of the reference subject. We report the metrics of Stable Diffusion with and without Pick-and-Draw in Tab.~\ref{tab::quantitative}. The subject fidelity is significantly improved at the cost of minor image-text fidelity, and this training-free approach demonstrates comparable overall performance to Textual Inversion. This may inspire further research on training-free single image text-to-image personalization.

\section{Conclusion}
In this paper, we propose Pick-and-Draw, a training-free semantic guidance approach for text-to-image personalization. We point out the prevalent overfitting issue of current methods: (1) they tend to memorize the few reference samples and struggles to generate diverse poses, views and backgrounds of the subject; (2) they require careful hyperparameter tuning to achieve delicate results. To this end, we propose appearance picking guidance and layout drawing guidance to boost performance for any personalized models with a single reference image. Qualitative and quantitative experiments demonstrate that Pick-and-Draw consistently improves identity consistency and generative diversity, pushing the trade-off between subject fidelity and image-text fidelity to a new Pareto frontier.

\clearpage
\bibliographystyle{named}
\bibliography{ijcai24}

\clearpage

\appendix

\section{Details of DreamBench Dataset}
DreamBench is a dataset collected by Ruiz \etal~\shortcite{ruiz2023dreambooth} for text-to-image personalization performance evaluation.  It consists of 30 subjects, including unique objects and pets such as backpacks, stuffed animals, dogs, cats, toys, etc. The subjects are separated into two categories, where 21 are objects and 9 are live subjects/pets. 25 text prompts are collected for testing generalization ability, including recontextualization and property modification. The evaluation suite requires generating 4 images for each subject and each prompt, amounting to a total of 3000 images.

\section{Details of Evaluation Metrics}
We follow DreamBooth and employ evaluation metrics of DINO, CLIP-I, and CLIP-T scores. DINO and CLIP-I score are utilized to evaluate subject fidelity, while CLIP-T score is utilized to evaluate image-text fidelity. The DINO score is the average pairwise cosine similarity between the ViT-S/16 DINO embeddings of the generated and real images. The CLIP-I score is the average distance of pairwise CLIP ViT-B/32 image embeddings of the generated and real images. The DINO score is considered a preferred metric for measuring subject fidelity due to its sensitivity in capturing variations among subjects within the same class. These two combined reflect identity consistency. Lastly, the CLIP-T score represents the average cosine similarity between the CLIP embeddings of text prompt and image. CLIP-I measures image-text alignment among various contexts, thereby reflecting model's generative diversity.

\section{Details of Baselines}
\noindent\textbf{Textual Inversion}~\cite{gal2022image} is a fine-tuning based method which optimizes a placeholder embedding of the diffusion text encoder, so as to invert the subject into the diffusion text space. It requires 2000 $\sim$ 3000 training steps for learning a new subject and we report results using 2500 steps across all instances in the experiments on DreamBench.
\\

\noindent\textbf{DreamBooth}~\cite{ruiz2023dreambooth} is a fine-tuning based method which optimizes the parameters of the whole diffusion UNet for image personalization. It learns to bind the specified subject with a rare text token via a reconstruction loss. It further utilizes a class prior preservation loss to avoid language drift and reduced generative diversity, but at the cost of worsening identity consistency. It requires around 400 $\sim$ 800 steps in general and we report the results using 600 steps in the experiments on DreamBench.
\\

\noindent\textbf{BLIP-Diffusion}~\cite{li2023blip} is an encoder-based method which pre-trains a multimodal encoder following BLIP-2 to produce visual representation aligned with the text. It exhibits zero-shot capability but needs further fine-tuning to achieve better performance. It requires 40 $\sim$ 120 steps for different subject and we report the results using 80 steps in the experiments on DreamBench.

\section{More Ablation and Analysis}
\begin{figure*}[htb]
    \begin{center}
        \includegraphics[width=0.68\linewidth]{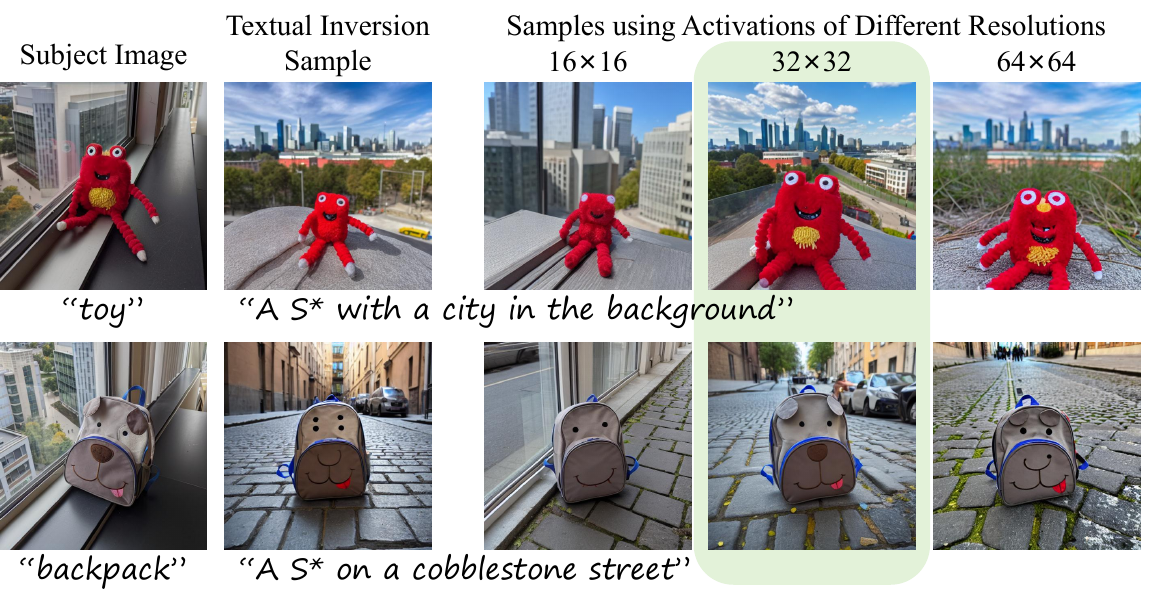}
    \end{center}
    \vspace{-2mm}
    \caption{Ablation study for visualizing different activation selection strategy for the appearance picking guidance, conducted on Textual Inversion. The best selection strategy is marked in green.}
    \label{fig:ablation_activation}
\end{figure*}

\noindent\textbf{Visualization of activation selection.}
In Fig.~\ref{fig:ablation_activation} we visualize different activation selection strategies for the appearance picking guidance. We choose Textual Inversion as example baseline, and generate samples with Pick-and-Draw guidance using activations of resolution $16\times 16$, $32\times 32$ and $64\times 64$. Activations of resolution $16\times16$ contain mainly layout information, which is coarse-grained and may introduce unwanted layout leakage (\ie the window frame) to the guided samples ($3\textsuperscript{rd}$ column). Activations of resolution $64\times64$ focus on high-frequency details such as edges, which are not sufficient for local appearance transfer. The corresponding guided samples ($4\textsuperscript{th}$ column) fail to maintain identity consistency with the reference image. In comparison, activations of resolution $32\times32$ encode rich semantic information and focus on different salient regions of the subject, facilitating the local appearance transfer to achieve the best visual result in identity preservation. Therefore, the guided samples (last column) not only exhibit consistent appearances, but also eliminate the interference of background from reference images.
\\

\begin{figure*}[htb]
    \begin{center}
        \includegraphics[width=0.8\linewidth]{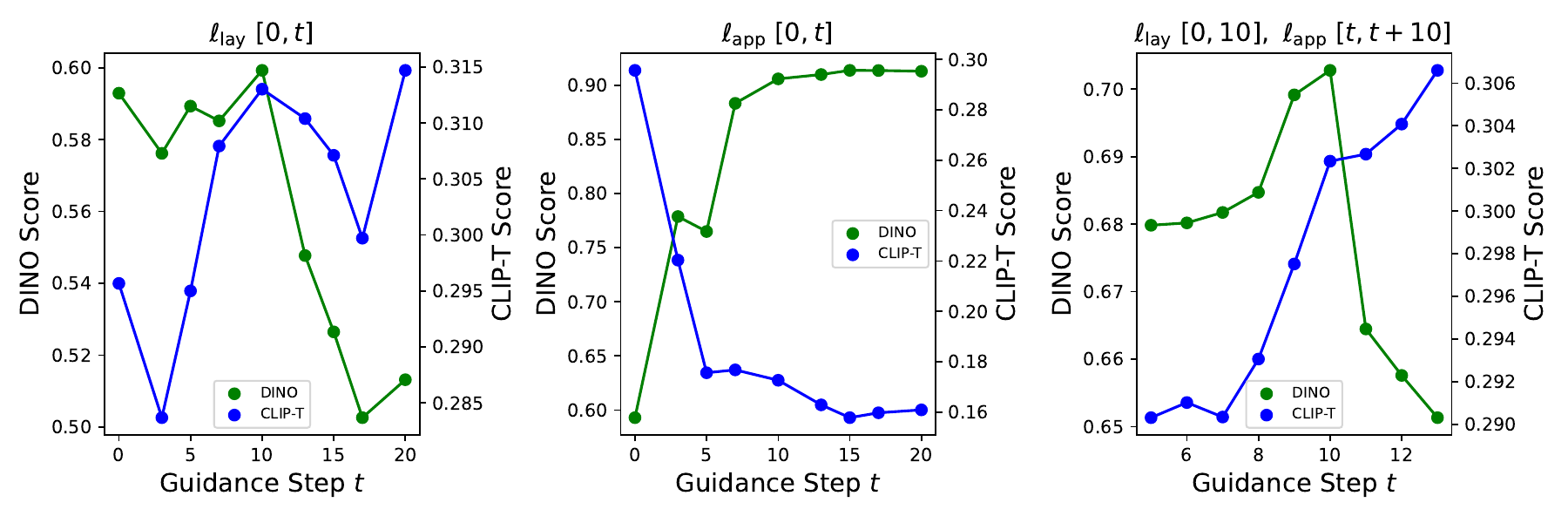}
    \end{center}
    \vspace{-2mm}
    \caption{Ablation study for guidance step selection. We conduct ablation experiments under three conditions: employing only the layout loss (\textit{left}), employing only the appearance loss (\textit{middle}), and employing both losses simultaneously (\textit{right}). The guidance steps of both losses are labeled.}
    \label{fig:ablation_timestep}
\end{figure*}

\noindent\textbf{Impact of guidance step selection.} We conduct ablation study on guidance steps for appearance-aware loss $\ell_{\text{app}}$ and layout-aware loss $\ell_{\text{lay}}$ on DreamBooth. Results are presented in Fig.~\ref{fig:ablation_timestep}. Since early denoising steps exert a significant impact on the generated object layout~\cite{hertz2022prompt,chefer2023attend}, we perform layout guidance solely (Fig.~\ref{fig:ablation_timestep} \textit{left}) from the very beginning and find that the optimal performance is achieved when stopping guidance at step 10. More layout guidance steps result in significant performance drop on DINO score. We perform appearance guidance (Fig.~\ref{fig:ablation_timestep} \textit{middle}) in a similar manner and find that 10 steps are sufficient for appearance transfer. Additionally, applying only appearance guidance leads to substantial decrease in the CLIP-T score, indicating severe overfitting problem. We then set the guidance schedule of $\ell_{\text{lay}}$ as $[0, 10]$ and fix the length of appearance guidance as 10 steps, and perform the two types of guidance simultaneously (Fig.~\ref{fig:ablation_timestep} \textit{right}). Scatter plot shows that starting the appearance guidance at step 10 achieves the best trade-off performance. 

Considering above, we set the range of guidance steps for layout guidance and appearance guidance as $[0, 10]$ and $[10, 20]$, respectively. The optimal setting is in line with intuition, where we initially employ layout-aware loss to constrain subject shape and background to ensure generative diversity and image-text fidelity, followed by the utilization of appearance loss to enforce object identity consistency and subject fidelity. We adhere to this setting throughout all other experiments.

\section{Failure Cases}
\begin{figure*}[htb]
    \begin{center}
        \includegraphics[width=0.82\linewidth]{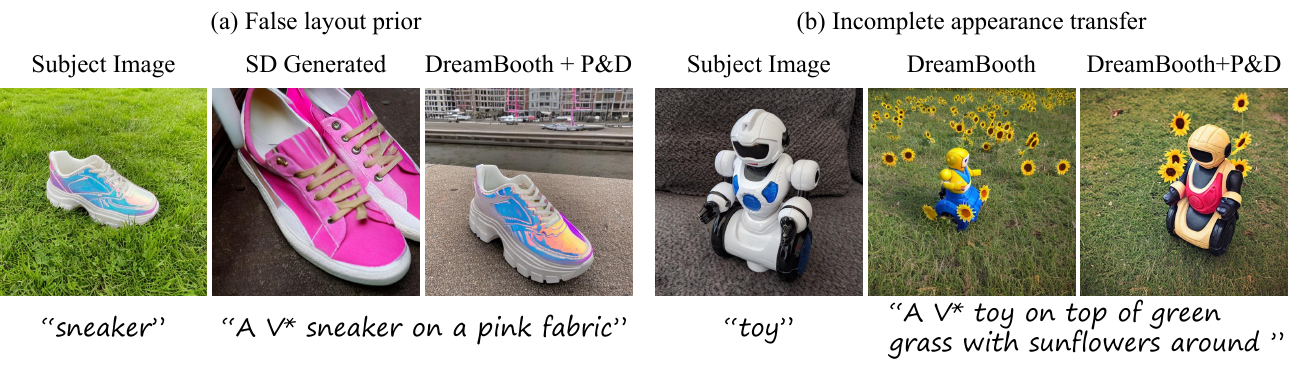}
    \end{center}
    \caption{Example failure generations. SD stands for Stable Diffusion and P\&D stands for our method Pick-and-Draw.}
    \label{fig:failure_case}
\end{figure*}

Pick-and-Draw fails to generate images aligned with text prompts if the template image by Stable Diffusion provides false layout prior. In addition, it may suffer from incomplete appearance transfer when the subjects generated by baseline model differ too much from the reference. We present two possible failure cases in Fig.~\ref{fig:failure_case}. 

\section{More Qualitative Results}
\begin{figure*}[htbp]
    \begin{center}
        \includegraphics[width=1.0\linewidth]{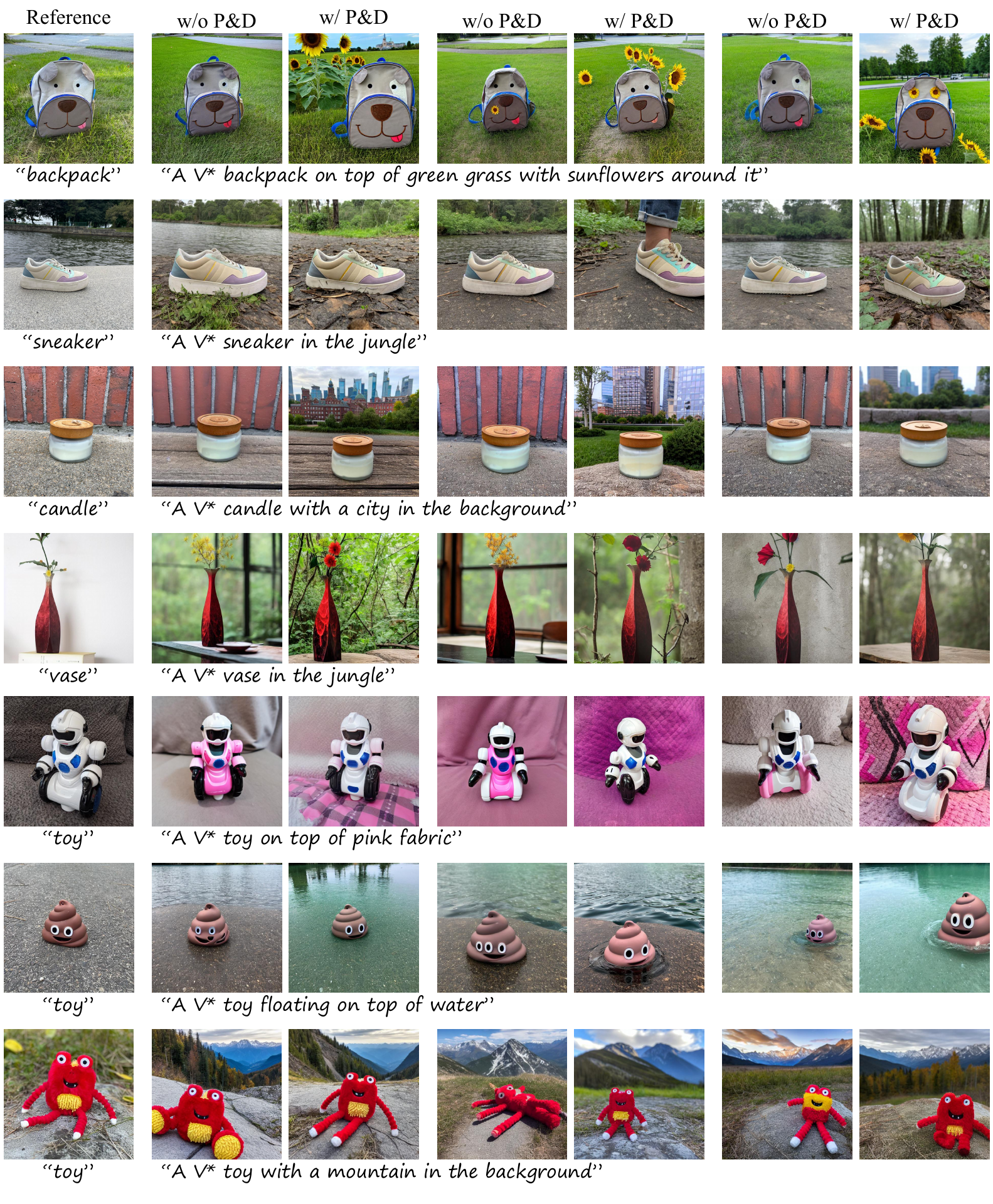}
    \end{center}
    \caption{More qualitative results on DreamBooth before and after applying Pick-and-Draw.}
    \label{fig:qualitative_db_suppl}
\end{figure*}

\noindent\textbf{Results on DreamBooth.} We provide more qualitative results in Fig.~\ref{fig:qualitative_db_suppl} to show the improvement of DreamBooth when equipped with Pick-and-Draw. Our method consistently improves performance of DreamBooth on both subject fidelity (row 5 $\sim$ 7) and image-text fidelity (row 1 $\sim$ 6).
\\

\begin{figure*}[h]
    \begin{center}
        \includegraphics[width=1.0\linewidth]{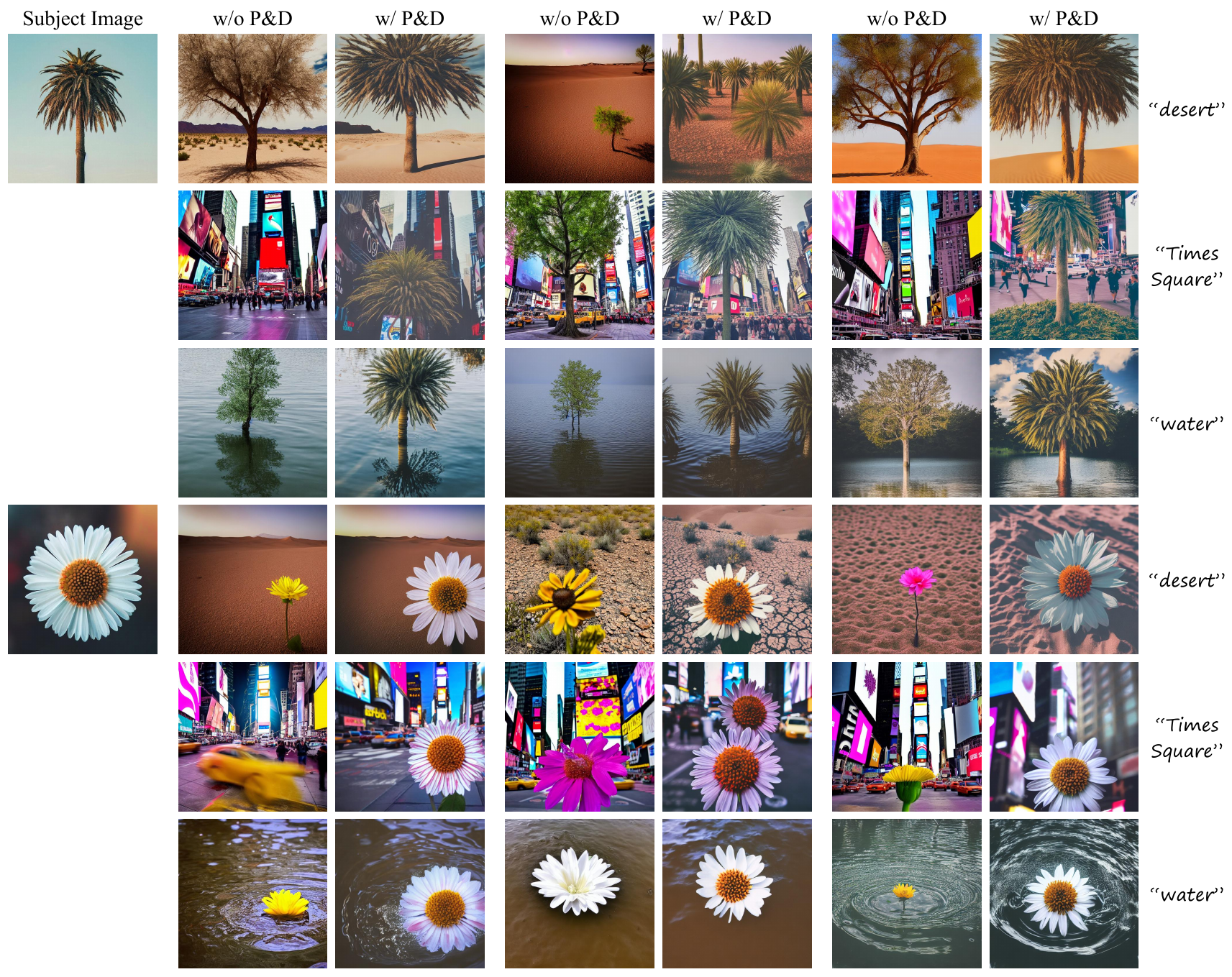}
    \end{center}
    \caption{More qualitative results on Vanilla Stable Diffusion before and after applying Pick-and-Draw.}
    \label{fig:qualitative_sd_suppl}
\end{figure*}

\noindent\textbf{Results on Vanilla Stable Diffusion.} We apply Pick-and-Draw to Vanilla Stable Diffusion for zero-shot text-to-image personalization. Visual results can be shown in Fig.~\ref{fig:qualitative_sd_suppl}.

\vspace{4mm}

\end{document}